\documentclass[letterpaper, 10 pt, conference]{ieeeconf}  

\IEEEoverridecommandlockouts                              

\usepackage{cite}
\usepackage{amsmath,amssymb,amsfonts}
\usepackage{algorithmic}
\usepackage{graphicx}
\usepackage{textcomp}
\usepackage{xcolor}
\usepackage{hyperref}
\usepackage{tabularx}
\usepackage{subfigure}
\usepackage{siunitx}

\title{\LARGE \bf
DRQN-based 3D Obstacle Avoidance with a Limited Field of View
}

\author{Yu'an Chen, Guangda Chen, Lifan Pan, Jun Ma, Yu Zhang, Yanyong Zhang, and Jianmin Ji$^*$
\thanks{
The work is partially supported by the National Key Research and Development Program of China (No. 2018AAA0100500), CAAI-Huawei MindSpore Open Fund, Anhui Provincial Development and Reform Commission 2020 New Energy Vehicle Industry Innovation Development Project "Key System Research and Vehicle Development for Mass Production Oriented Highly Autonomous Driving", and Key-Area Research and Development Program of Guangdong Province 2020B0909050001.
}
\thanks{School of Computer Science and Technology, University of Science and Technology of China, Hefei, 230026, China}%
\thanks{ $^*$ Corresponding authors. {\tt\small \{jianmin\}@ustc.edu.cn}}%
}%

\begin{document}

\maketitle
\thispagestyle{empty}
\pagestyle{empty}

\begin{abstract}

In this paper, we propose a map-based end-to-end DRL approach for three-dimensional (3D) obstacle avoidance in a partially observed environment, which is applied to achieve autonomous navigation for an indoor mobile robot using a depth camera with a narrow field of view. 
We first train a neural network with LSTM units in a 3D simulator of mobile robots to approximate the Q-value function in double DRQN. We also use a curriculum learning strategy to accelerate and stabilize the training process.
Then we deploy the trained model to a real robot to perform 3D obstacle avoidance in its navigation. 
We evaluate the proposed approach both in the simulated environment and on a robot in the real world. The experimental results show that the approach is efficient and easy to be deployed, and it performs well for 3D obstacle avoidance with a narrow observation angle, which outperforms other existing DRL-based models by 15.5\% on success rate.

\end{abstract}

\section{Introduction}\label{introduction}

Autonomous mobile robots have become increasingly popular in recent years, which are capable to provide services like delivery, cleaning, guidance, and so on. Safe navigation in complex and crowded environments is essential for these applications.
In this paper, we consider the navigation problem for indoor robots.
Within this well-researched domain, mainstream approaches consist of a global planner and a local planner. The global planner creates a trajectory or waypoints using a pre-built map of the environment. Then, the local planner follows the global path while avoiding collisions with static obstacles but also unexpected dynamic obstacles, which is one of the well-known challenges in robot navigation.

Most indoor mobile robots have been developed considering a two-dimensional (2D) planar environment, that relies on 2D range finders
to provide 2D maps of the obstacles that can be obtained in a plane horizontal to the ground. However, such 2D maps may lose 3D coordinate information of obstacles, like table corners, low steps, feet, etc., which affect the safety of navigation in office-like environments.

Recently, obstacle avoidance in a 3D dynamic environment with partial observation has received considerable attention. 
Marder-Eppstein et al.~\cite{marder2010office} used a couple of 3D laser range finders to accomplish safe navigation in a complex office environment. Wang et al.~\cite{wang2017autonomous} employed an RGB-D camera and a 2D laser to make their robot moving autonomously in an uneven and unstructured indoor environment. 
Most existing algorithms for 3D obstacle avoidance are extended from 2D methods like Dynamic Window Approach (DWA)~\cite{fox1997dynamic} and elastic bands~\cite{quinlan1993elastic}. 
These extensions aggravate their weaknesses of intensive computational demand and limited generalization to unseen environments, since they are sample-based and involve numerous parameters that need to be tuned manually.

To overcome these weaknesses, deep reinforcement learning (DRL) based control approaches were proposed ~\cite{chen2017socially,long2018towards,fan2018crowdmove,zhu2017target,wu2018building,choi2019deep}. These approaches learn neural networks that directly map sensor inputs to the velocities of the robot by interacting with the environment. 
Although DRL approaches have shown promising results in robot navigation, few works have been done for 3D obstacle avoidance without the use of expensive 3D LiDAR scanners.

In this paper, we propose an end-to-end DRL algorithm for 3D obstacle avoidance in partially observed office-like environments using a Kinect v1 depth sensor (RGB-D camera). 

Note that, this RGB-D camera provides depth information of the environment. 
Then we can convert the information into a costmap and apply a map-based DRL approach for 3D obstacle avoidance, which helps us to avoid multiple tough challenges common in visual navigation~\cite{zhu2017target}.
However, the main limitation of RGB-D cameras is their narrow Field of View (FOV).
Here, Kinect v1 has a horizontal FOV of 57-degree and a vertical FOV of 43-degree.
To handle the limited FOV of the depth camera, we use double Deep Recurrent Q-Learning (DRQN)~\cite{hausknecht2015deep} to infer the drivable area from consecutive probabilistic occupancy grid maps (costmaps)
generated from sensor data. 
We also use a curriculum learning strategy to accelerate and stabilize the training process.
We consider local costmaps as inputs of the trained network. Then it is easier to deploy the trained model to a real robot to perform 3D obstacle avoidance in the real world.
We demonstrate and evaluate the approach in both simulation and real-world experiments. The experimental results show that our approach is efficient and easy to be deployed, and it performs well for 3D obstacle avoidance with a narrow observation angle, which outperforms other existing DRL-based models in numerous indicators.

The main contributions of our work are the followings:
\begin{itemize}
	\item We propose an end-to-end DRL method for 3D obstacle avoidance in partially observed environments.
	\item We verify the feasibility of using a depth camera with a narrow FOV for 3D obstacle avoidance by DRL.
	\item We show that the trained model is easy to be deployed to a real robot by considering local costmaps as inputs of the neural network.
\end{itemize}


\section{Related Work}\label{Related Works}

Traditional robot navigation approaches~\cite{mohanan2018survey,fox1997dynamic,quinlan1993elastic} normally depend on several human-engineered hyperparameters and rules that are not likely to be satisfied in practice. Then these approaches might fail in some cases due to the obvious sensitivity of these hyperparameters and rules.

To overcome the limitation, DRL-based approaches are widely studied. In DRL, the robot interacts with the environment and learns from numerous trials with corresponding rewards instead of labeled data to map sensor inputs directly to its velocities. Based on the differences in input data, existing DRL-based navigation approaches can be roughly divided into two categories: methods with agent-level inputs and methods with sensor-level inputs. 
In specific, methods with agent-level inputs require expensive motion data of other robots, like velocities, accelerations, and paths, for the states of the DRL model and 
methods with sensor-level inputs consider sensor inputs as the states of the DRL model.

As a representative of methods with agent-level inputs, Chen et al.~\cite{chen2017socially} trained an agent-level collision avoidance policy using DRL, which maps an agent's state and its neighbors' states to collision-free action. However, it demands perfect sensing, which makes the whole system less robust to noise and difficult to be deployed to a real robot.

For methods with sensor-level inputs, various types of sensor data are considered in DRL-based navigation, including 2D Lidar sensor data~\cite{fan2018crowdmove,long2018towards,chen2020robot}, color images~\cite{zhu2017target}, depth images~\cite{zhang2017deep}, and 3D point clouds~\cite{choi2019deep}.
In specific, Fan et al.~\cite{fan2018crowdmove} proposed a convolutional neural network (CNN) with the last three consecutive frames of laser range as inputs and control commands as outputs.
Similarly, Long et al.~\cite{long2018towards} and Chen et al.~ \cite{chen2020robot} directly derive the control commands from laser range sensors through neural networks. These approaches have been deployed to real robots. However, expensive LiDAR scanners are required to provide a wide FOV (\ang{180} in \cite{fan2018crowdmove,long2018towards,chen2020robot}) sensing.

DRL-based navigation with a 2D laser can be easily deployed to a real robot, as the difference between sensing data of a simulated 2D laser and a real one is limited.
However, 2D sensing data is not enough to describe complex 3D scenarios. On the contrary, vision sensors can provide 3D sensing information. Zhu et al.~\cite{zhu2017target} proposed a DRL framework for target-driven visual navigation using color images. And Zhang et al.~\cite{zhang2017deep} proposed a successor-feature-based DRL algorithm that can quickly adapt to new navigation tasks with inputs of depth images.
However, for both RGB and RGB-D images there are large differences between the images generated in simulated environments and the ones generated in the real-world, which makes it difficult to make full use of the simulator and deploy the trained model from a simulator to a real robot. 
On the other hand, an RGB-D camera can also be considered as a depth camera, which provides 3D point clouds of the environment. 
Compared to vision inputs, DRL-based navigation with 3D point clouds is easier to be deployed to a real robot. 
However, the FOV of a depth camera is usually narrow, which severely degrades the performance of DRL methods. To overcome the problem, Choi et al.~\cite{choi2019deep} proposed an LSTM based model with Local-Map Critic (LSTM-LMC) for DRL-based navigation with 3D point clouds generated by a depth camera. 
However, the depth camera in this paper is much narrower than theirs, which greatly increases the difficulty of the navigation problem. 
Lobos-Tsunekawa et al.~\cite{lobos2020point} used Simultaneous Localization and Mapping (SLAM) to directly integrate the information of the environment from previous time steps.
Note that, their navigation policy only needs to choose an action from three discrete actions, which limits their applications in various robots in the real world.

The approach in this paper stems from our previous work ~\cite{chen2020robot} and extends 2D obstacle avoidance to 3D obstacle avoidance. As same as the previous work, local costmaps are considered as inputs of the neural network, which helps us to extend the sensing data from a 2D laser scanner to a depth camera. The main differences are the perception sensor used and the corresponding improvements to the network structure.
Inspired by the approach in~\cite{choi2019deep}, we replace the expensive laser range finder with an affordable depth camera (Kinect v1) to implement obstacle avoidance in the method of DRL. In contrast to their work, only sensor-level input data are needed to generate our local costmaps while agent-level inputs such as types and speeds of obstacles are not required, which reduces the computational load of the network and makes it easier to deploy the trained model to a real robot.
Notice that, the FOV of our depth camera is narrower than the camera used in~\cite{choi2019deep}
. 
In this paper, we focus on applying DRL-based navigation for 3D obstacle avoidance in a partially observed indoor environment for a real robot with only a Kinect sensor.

\section{Problem Formulation}\label{Problem Formulation}

The obstacle avoidance problem considered here be formulated as a sequential decision-making problem with partial observation, due to the limited FOV of the depth camera.

At each time step $ t $, the robot receives a partial observation $ o_{t} $ and computes a collision-free control command that navigates it to move from the starting position $p_{0}$ to the goal position $p_{g}$. The observation $ o_{t} $ drawn from
the probability distribution w.r.t. the underlying system state $ s_{t}$, i.e., $o_{t} \sim O\left(s_{t}\right) $, only provides partial information about the state $ s_{t} $. 
The observation of each time step can be divided into three parts: 
\[
o_{t} = \left[M_{t},o_{t}^{g},o_{t}^{v}\right],
\]
where $ M_{t} $ is a local costmap contains information of raw depth point cloud measurements about its ahead environment, $ o_{t}^{g} $ stands for its relative goal position, and $ o_{t}^{v} $ refers to its current velocity. According to the observation $ o_{t} $, the robot computes an action, $a_{t}$, sampled from a stochastic policy $ \pi $ learned from past experiences:
\[
a_{t} \sim \pi_{\theta}\left(a_{t} | o_{t}\right), 
\]
where $ \theta $ indicates the parameters of the policy. The computed action $ a_{t} $ consists of the directed linear velocity $ v_{t} $ and angular velocity $ \omega_{t} $ for the robot. 
A sequence of actions $\langle a_0, a_1, \ldots, a_{t_g} \rangle$
can be considered as a trajectory that navigates the robot to move from its starting position $ p_{0} $ to its goal position $ p_{g} $, where $ t_{g} $ is the last time step of the sequence. 
Our goal is to find the optimal policy that minimizes the expectation of the mean arrival time while avoiding all obstacles, which is defined as:
\[
\begin{aligned} 
\underset{\pi_{\theta}}{\operatorname{argmin}} {\mathbb{E}\left[\,t_{g} |\right.}{\mathbf{v}_{t}} &=\pi_{\theta}\left(a_{t}|{o}_{t}\right), \\
{p}_{t} &={p}_{t-1}+{\mathbf{v}_{t}} \cdot \Delta t, \\
\forall k &\left.\in[1, N]:\left\|{p}_{t}-\left({p}_{o b s}\right)_{k}\right\|>R\,\right],
\end{aligned}
\]
where ${p}_{obs}$ is the position of obstacles, $ R $ is the robot radius.

\section{Approach}\label{approach}


\subsection{Environment Representation}

An accurate and compact representation of the environment is crucial for a navigation system as it captures all information needed for the navigation component to general collision-free and optimal motions for the robot.

\subsubsection{Point Cloud Filter Pipeline}

The original 3D point cloud contains outliers and redundant information. In the point cloud filter pipeline, we first down-sample the point cloud, then remove outliers and refine the result by eliminating the ceiling.

Firstly, we perform a down-sampling procedure to reduce the resolution and the computational costs of the point cloud. 
In specific, 
we convert the point cloud to 3D voxel grids with the side length of 5cm and use centers of all voxels to down-sample the point cloud.

Secondly, we apply a statistical filter
to remove outliers to reduce the noise in the point cloud. 
The sparse outlier filter is based on the computation of the distribution of point to neighbor distances in the input dataset. For each point, we compute the mean distance from it to all its 50 neighbors. By assuming that the resulted distribution is Gaussian with a mean and a standard deviation, all points whose mean distances are outside an interval defined by the global distances mean and standard deviation can be considered as outliers and trimmed from the dataset.

Lastly, we refine the result and refrain from regarding the floor and ceiling as obstacles. In specific, we remove all points that are higher than the height of our robot (1.35 m). 

After being processed by the point cloud filter pipeline, the remaining points in the cloud would be further utilized to construct the costmap.

\begin{figure}[t]
	\centering
	\subfigure[]{
		\label{fig1:subfig:a} 
		\includegraphics[width=3.5cm]{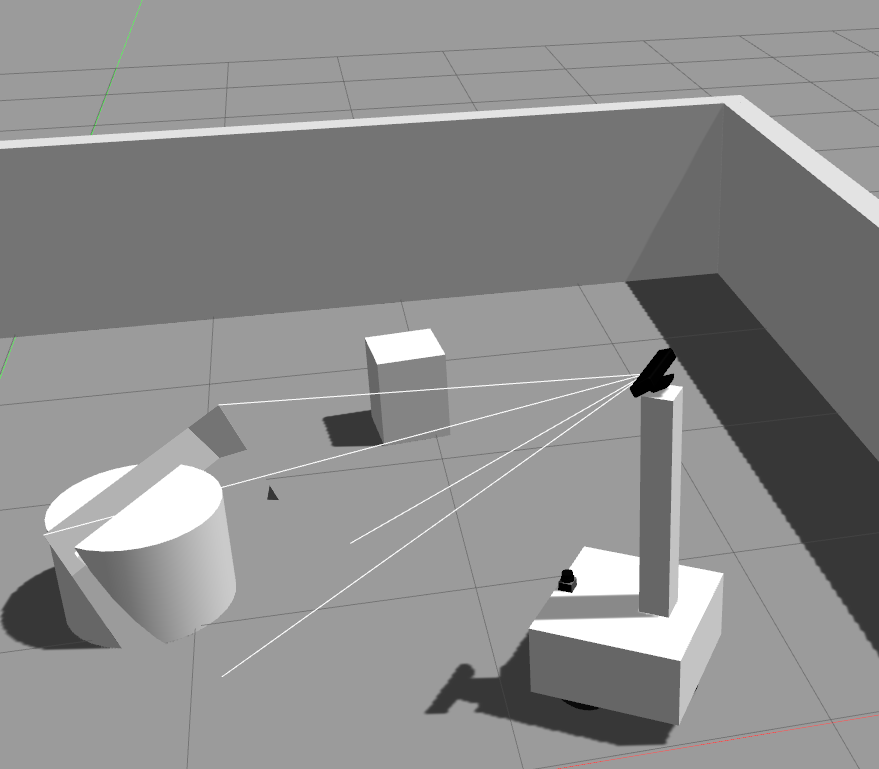}
	}
	\subfigure[]{
		\label{fig1:subfig:b}
		\includegraphics[width=3.5cm]{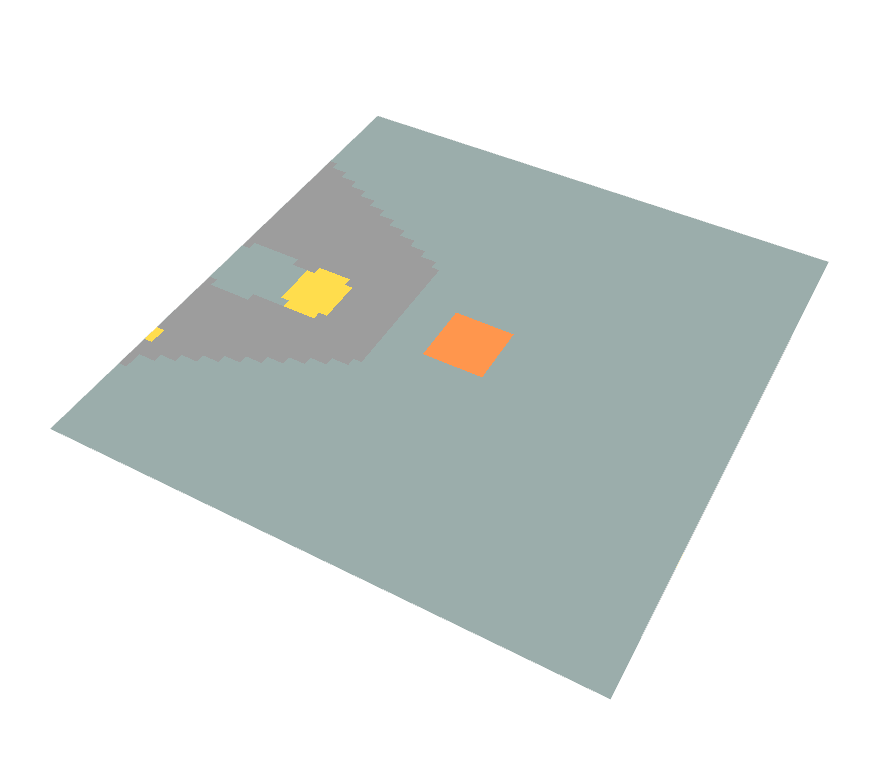}
	}
	\caption{(a) A Gazebo training environment and (b) the corresponding local costmap displayed in Rviz.} 
	\label{fig1:subfig}
\end{figure}

\begin{figure*}[t]
	\centering{\includegraphics[width=15cm,height=4.5cm]{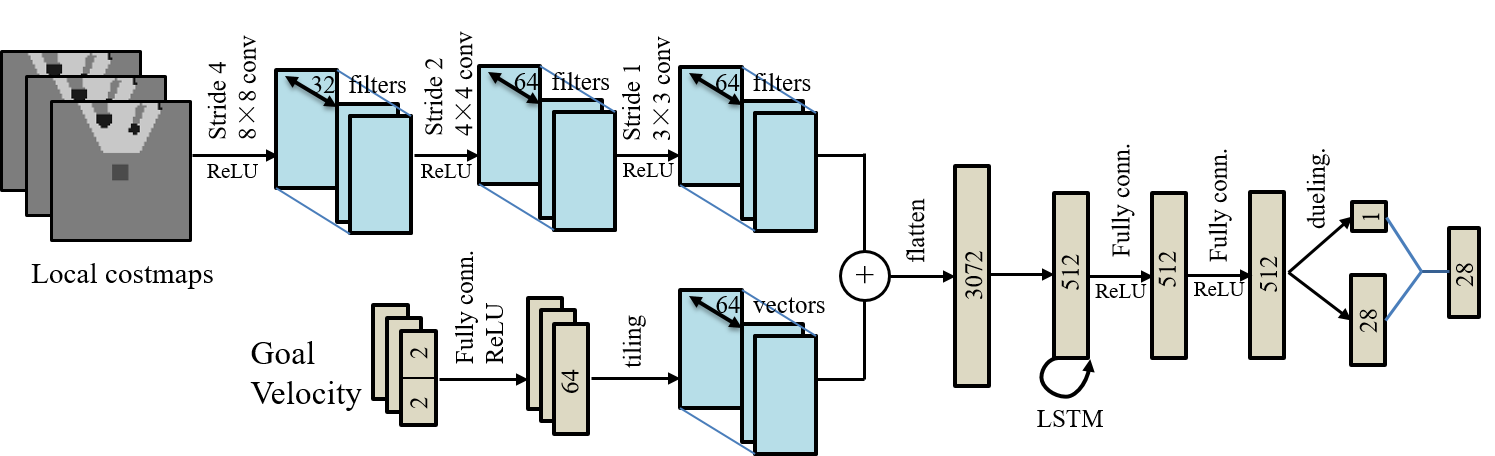}}
	\caption{The architecture of our double DRQN network. This network takes
		three consecutive frames of local costmaps, the goal position and the robot's velocity as inputs, and outputs extended Q-values for 28 different actions.}
	\label{fig2}
\end{figure*}

\subsubsection{Costmap Generator}

We use layered costmaps~\cite{lu2014layered} to represent the environment
perceived by multiple sensors. 
In this paper, as long as the robot is moving on flat ground,
we can transform the processed 3D point cloud into the bird's-eye view 2D costmap.

In specific, we initialize the costmap with unknown values and update it when a new sensor reading is received, which makes the costmap maintain an up-to-date view of the robot's local environment. 
Although the costmap is a 2D structure, it still can efficiently specify the compressed 3D information of the point cloud. 
Note that, the processed points could be divided into vertical columns. Then the costmap generator can project these columns into corresponding values for grids of the costmap. The points lower than 5 cm will be regarded as free space and other points are considered as obstacles. 

We can also add the robot configuration (shape) into the costmap, by which the neural network can recognize the colliding situations, i.e., the overlap situations of the robot configuration and objects in the costmap.
Moreover, we convert the costmap to a corresponding image to ease the setup of the network. Without loss of generality, we identity a costmap with its corresponding image in the paper. 
Fig.~\ref{fig1:subfig:b} shows an example of the generated costmap that covers a 6 m$ \times $6 m area around the robot with a resolution of 10 cm.

\subsection{Double DRQN}

The partially observable sequential decision-making problem
defined in Section~\ref{Problem Formulation} can be considered as a Partially
Observable Markov Decision Process (POMDP) problem~\cite{boyen2013tractable}, which can be solved using reinforcement learning.
In specific, a POMDP problem consists of 6 tuples $\langle S,A,P,R,\Omega,O \rangle$, where $S$ is the state space, $A$ is the action space, $P$ is the transition probability function, $R$ is the reward function, $\Omega$ is observation space with $o\in \Omega$, and $O$ is observation probability function.

The objective of reinforcement learning ~\cite{sutton1998introduction} is to learn the policy of the agent, $\pi_{\theta}(a,o) = p(a\mid o)$, that maximizes the discounted return
\[
G=\sum_{t=0}^{\infty} \gamma^{t} \mathbb{E}\left[r\left(s_{t}, a_{t}\right)\right],
\]
where $ \gamma \in \left[0,1\right] $ is the  discount factor for future rewards.

In our previous work~\cite{chen2020robot}, Deep Q-Learning Network (DQN)~\cite{osband2016deep} is used for 2D obstacle avoidance. In DQN, the superiority of a policy $\pi$ is evaluated by the Q-value (action-value) function as:
\[
Q^{\pi}(s, a)=\mathbb{E}^{\pi}\left[\sum_{t=0}^{\infty} \gamma^{t} R\left(s_{t}, a_{t}\right) \mid s_{0}=s, a_{0}=a\right].
\]

However, DQN assumes full observation of the environment, which has no explicit mechanisms to decipher hidden states from observations in POMDP. 
Then, using DQN to estimate a Q-value from observations would go wrong, i.e., generally $Q\left(o,a\mid \theta\right) \neq Q\left(s,a\mid \theta\right)$.

To handle these POMDP problems, Hausknecht et al.~\cite{hausknecht2015deep} introduced Deep Recurrent Q-Networks (DRQN).
In DRQN, the Q-value $Q\left(s_{t},a_{t}\right)$ is no longer considered, but 
$Q\left(o_{t},h_{t-1},a_{t}\right)$ is used, where $ h_{t} $ is an extra input generated by the network at each time step $t$, which reflects the agent's current belief of hidden states.
Recurrent neural network units like LSTM~\cite{hochreiter1997long} can be placed in a DQN network for such a purpose, where $h_{t}= \text{\it LSTM}(h_{t-1},o_{t})$ and the extended Q-value $Q\left(o_{t},h_{t-1},a_{t}\right)$ is estimated by the network.

Some tricks that are effective for DQN can also be used to improve the performance of DRQN. In specific, inspired by Double Q-learning~\cite{van2016deep}, we use a double network for the extended Q-value function to eliminate the overestimation bias.
We also apply Dueling networks~\cite{wang2015dueling} for a more normalized estimation of extended Q-values.
We specify the details of our DRQN-based approach in the following. 

\subsubsection{Observation Space}

As mentioned in Section~\ref{Problem Formulation}, the observation $ o_{t} $ consists of the generated costmap $ M_{t} $, the relative goal position $ g_{t} $
and the robot's current velocities $ {\mathbf v}_{t} $. 
In specific, $M_t$ specifies the costmap image generated from the point cloud data from the depth camera with \ang{57} FOV. 
$ g_{t} $ is a 2D vector that specifies the goal position in the corresponding polar coordinates w.r.t. the robot's current position.
$\mathbf{v}_t$ specifies the linear and angular velocities of the robot at time~$t$.

\subsubsection{Action Space}

In this paper, we consider a two-dimensional action space composing linear velocity and angular velocity. 
We discrete the action space into 28 different actions, i.e., 4 values for the linear velocity, $v\in \{0.0,\, 0.2,\, 0.4,\, 0.6\} $, and 7 values for the angular velocity, $\omega \in \{-0.9,\, -0.6,\, -0.3, \, 0.0, \, 0.3,\, 0.6,\, 0.9 \}$, where an action is a pair $(v, \omega)$.
Note that, $v<0.0$ is not allowed, i.e., moving backward is not allowed, due to lack of rear sensors.

\subsubsection{Reward Function}

The reward $ r $ consists of four terms as follows :
\[
r = r_{goal}+r_{collision}+r_{safety}+r_{step}.
\]

In particular, 
$ r_{goal}$ specifies the penalty when the robot goes far from its goal. 
$r_{goal}=500$ when the robot reaches its goal.
We define $r_{goal}$ as:
\[
r_{goal}=\left\{\begin{array}{ll}
500 & \hspace{-.3cm} \text { if }\left\|{p}_{t}-p_{g}\right\|<0.3, \\
\varepsilon_{g}\left(\left\|{p}_{t-1}-p_{g}\right\|-\left\|{p}_{t}-p_{g}\right\|\right) & \text { otherwise.}
\end{array}\right.
\]
where $\varepsilon_g$ is the hyper parameter that controls the penalty amount. We set $\varepsilon_g = 200$.

$r_{collision}$ specifies the penalty when the robot encounters a collision. We define $r_{collision}$ as:
\[
r_{collision}=\left\{\begin{array}{ll}
-500 & \text { if the robot encounters a collsion}, \\
0 & \text { otherwise.}
\end{array}\right.
\]

$ r_{safety} $ specifies the penalty when the robot gets closer to obstacles.
We define $r_{safety}$ as:
\[
r_{safety} = \varepsilon_{s}\left(min_{d_{t-1}}-min_{d_{t}}\right),
\]
where $min_{d_{t}}$ denotes the distance between the robot and the closest obstacle boundary at time~$t$, and $\varepsilon_{s}$ is the hyper parameter that controls this penalty amount. We set $\varepsilon_{s} = -100$.

At last, we apply a small negative penalty for each time step, i.e., $r_{step}= -5$, to encourage short paths.

\subsubsection{Network Architecture}

Our network architecture for double DRQN is specified in Fig.~\ref{fig2}. 
The inputs of the network consist of three parts, i.e., three consecutive frames of local costmaps with $60\times60 $ gray pixels, the goal position, and the robot's velocity. The values of the goal position and the robot's velocity are combined into a four-dimensional vector. 

The network first produces feature maps with 64 channels for the costmaps using three convolutional layers\footnote{More details can be found in~\cite{hausknecht2015deep}.}.
The network also projects the goal position and the robot's velocity to vectors with 64 channels. Then the network combines these feature maps, flattens the combination, and feeds the result into the LSTM unit. At last, the network applies two fully connected layers and a dueling network structure to produce extended Q-values of these 28 discrete actions.

\subsection{Curriculum Learning}\label{curr}

Curriculum learning~\cite{bengio2009curriculum} has recently been shown as an efficient strategy that can help find better local minima and accelerate the training process.
The main idea of curriculum learning is to decompose a hard learning task into several simple ones, start with the simplest task, and level up the difficulty gradually.

In this work, we use Gazebo~\cite{koenig2004design} as our 3D simulation environment to train the robot for obstacle avoidance. 
In training mode, the environment will randomly choose locations for obstacles, the starting and target points of the robot. An example of such an environment is shown in Fig.~\ref{fig1:subfig:a}.
In the training process, we gradually increase the number of obstacles and the distance from the starting position to the target position to level up the complexity of the obstacle avoidance problem.
We level up the difficulty only when the robot has achieved a considerable success rate in current environments. 

\begin{figure}[t]
	\centering
	\subfigure[{A random scenario}]{
		\label{fig3:subfig:a}
		\includegraphics[width=2.3cm,height=2.3cm]{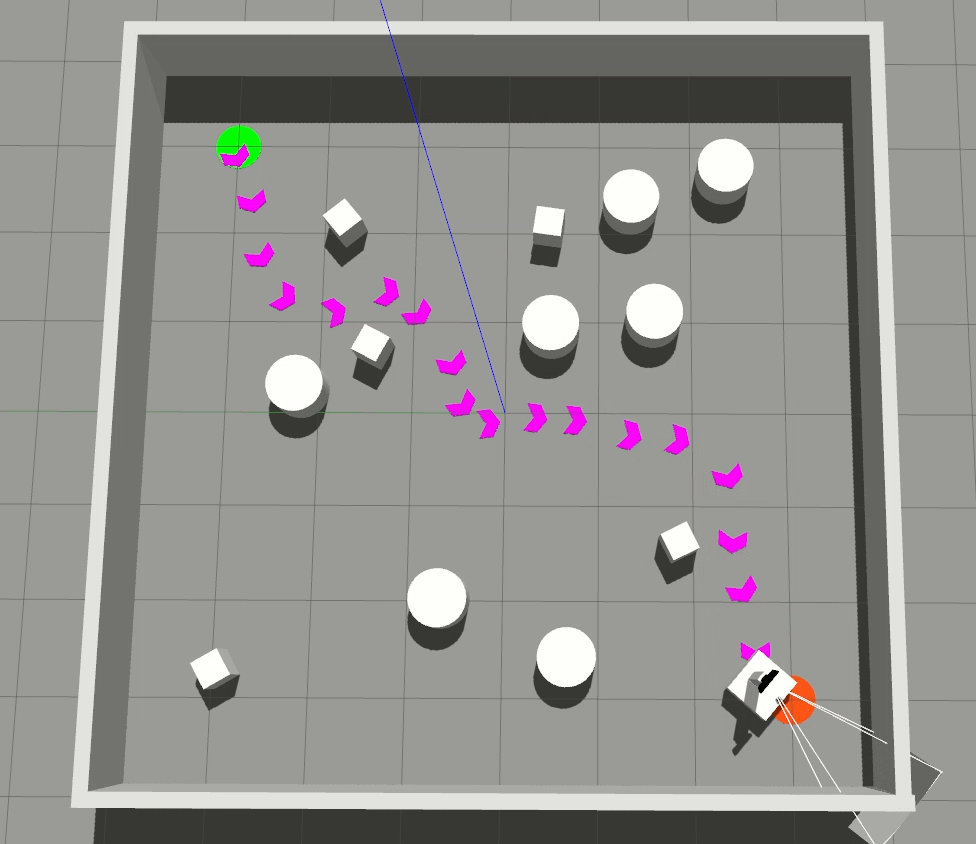}
	}
	\quad
	\subfigure[{An office-like scenario}]{
		\label{fig3:subfig:b} 
		\includegraphics[width=2.3cm,height=2.3cm]{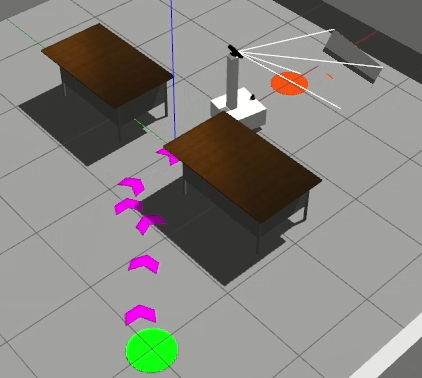}
	}
	\quad
	\subfigure[{ A coffee-like scenario}]{
		\label{fig3:subfig:c}
		\includegraphics[width=2.3cm,height=2.3cm]{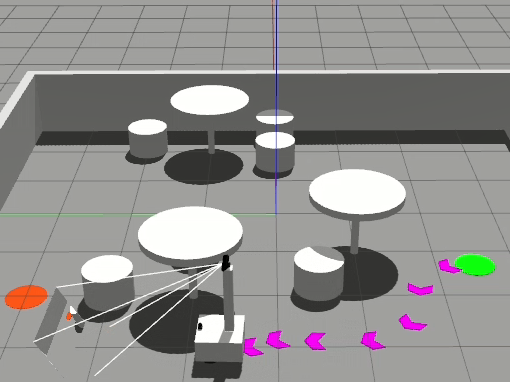}
	}
	\caption{Three testing scenarios used in our simulation experiments, the starting point is marked by a green dot, the target point is marked by a red dot, and the robot's trajectory is marked by purple arrows.} 
	\label{fig3:subfig}
\end{figure}

\section{Experiment}\label{experiment}


\begin{table}[b]
	\caption{Hyper Parameters}  
	\begin{center} 
		\begin{tabular}{ll}
			\hline\noalign{\smallskip}  
			Parameter & Value \\  
			\hline\noalign{\smallskip}  
			learning rate  & $ 1\times10^{-3} $ \\  
			discount factor & $ 0.97 $\\  
			replay buffer size & $ 2\times10^{5} $ \\
			minibatch size & $ 1024 $ \\
			image size & $ 60\times60 $ \\
			episode length & $200$ \\
			initial exploration & $1$ \\
			final exploration & $0.1$ \\
			LSTM unroll & $3$ \\
			LSTM initializer & $orthogonal$ \\
			\hline\noalign{\smallskip} 
		\end{tabular}
		\label{table1}  
	\end{center} 
\end{table}

\subsection{Implementation Details}

We implemented a simulated differential robot with a Kinect camera (depth camera with 57-degree horizontal FOV) in Gazebo, as illustrated in Fig.~\ref{fig1:subfig:a}.
Note that, the vertical FOV of the Kinect camera is limited to 43 degrees and the depth range of the Kinect is 0.8 - 4 m. Then Kinect cannot detect obstacles close to the robot. 
As illustrated in Fig.~\ref{fig1:subfig:b}, we specify the area in the near front of the robot as an unknown space, which increases the difficulty for obstacle avoidance.


We trained our DRQN-based agent with the hyperparameters listed in Table~\ref{table1}. The deep recurrent Q-network is implemented in TensorFlow and trained with the Adam optimizer~\cite{kingma2014adam}. A computer with an i9-9900k CPU and an NVIDIA GeForce 2080 Ti GPU is used for the training. It takes around 20 hours to train the network for good performance in testing environments. 

Besides our curricular DRQN-based approach (denoted by curricular DRQN), we also implemented other DRL-based approaches, i.e., curricular 3-frame DQN, the DQN-based approach introduced in~\cite{chen2020robot} using curriculum learning and taking three consecutive frames of local costmaps as inputs, and 3-frame PPO, an improved version of the PPO-based approach proposed in~\cite{long2018towards} also taking three consecutive frames of local costmaps instead of 2D laser data as inputs and
using 2D convolutions instead of 1D convolutions in the network.
Note that, curriculum learning is not applied in the original version of the PPO-based approach in~\cite{long2018towards}. We then implemented an extended version of it by adding curriculum learning. However, this extended version does not perform better than the original one. On the other hand, the improved version that replaces laser data and 1D convolutions with costmaps and 2D convolutions performs much better.

\subsection{Simulation Experiments}\label{slim}

\subsubsection{Evaluation Metrics}

We report four metrics to evaluate the performance of DRL-based approaches for obstacle avoidance as the following:
\begin{itemize}
	\item \textbf{Success Rate (SR)}:  the ratio of the episodes that end with the robot reaching its target without any collision.
	\item \textbf{Expected Return (ER)}: the average of the sum of rewards of all episodes.
	\item \textbf{Reach Time (RT)}: the average time for the robot to successfully reach the target without any collision.
	\item \textbf{Average Angular Velocity Change (AAVC)}: the average absolute value of the difference of angular velocities between two consecutive time steps, by which we can evaluate the smoothness of the trajectory.
\end{itemize}

\subsubsection{Training and Testing Scenarios}

As illustrated in Fig.~\ref{fig3:subfig:a}, we train the robot for obstacle avoidance in random scenarios of the simulation environment.
As mentioned in Section~\ref{curr}, we trained the two curricular approaches in the same training procedure.

Besides random scenarios, we also construct two new testing scenarios to evaluate the performance of trained models for unseen scenarios, i.e., office-like scenarios and coffee-like scenarios as illustrated in Fig.~\ref{fig3:subfig:b} and Fig.~\ref{fig3:subfig:c}.
In specific, an office-like scenario would require the robot to pass through the narrow passage between two desks and a coffee-like scenario would be filled with round chairs and round tables with one foot, where 3D obstacle avoidance is required. 
Note that, both office-like and coffee-like scenarios are challenging for most 2D obstacle avoidance approaches. 

\begin{table}[b]
	\caption{Metrics for three DRL-based approaches in random scenarios}
	\begin{center}
		\begin{tabular}{lllll}
			\hline\noalign{\smallskip}  
			Method & SR & ER & RT & AAVC \\  
			\hline\noalign{\smallskip}  
			Prior work  & 76.5 & 445 & 7.94 & 0.56 \\  
			PPO & 57.5 & 404 & \textbf{5.88} & 0.49 \\  
			Ours & \textbf{92} & \textbf{715} & 6.78 & \textbf{0.43} \\
			\hline\noalign{\smallskip} 
		\end{tabular}  
		\label{table2}  
	\end{center}
\end{table}

\subsubsection{Comparative Experiments}

Now we compare the performance of three DRL-based approaches for 3D obstacle avoidance, i.e., ours (curricular DRQN), the prior work (curricular 3-frame DQN), and 3-frame PPO. 

\begin{figure}[t]
	\centering
	\subfigure[]{
		\label{fig4:subfig:a} 
		\includegraphics[width=0.46\linewidth]{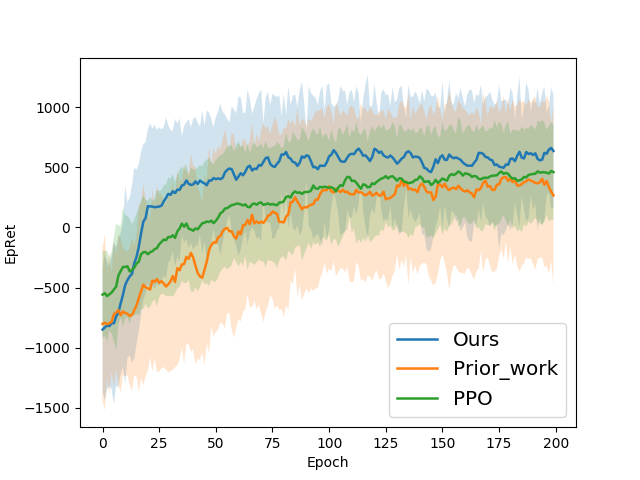}
	}
	\subfigure[]{
		\label{fig4:subfig:b}
		\includegraphics[width=0.46\linewidth]{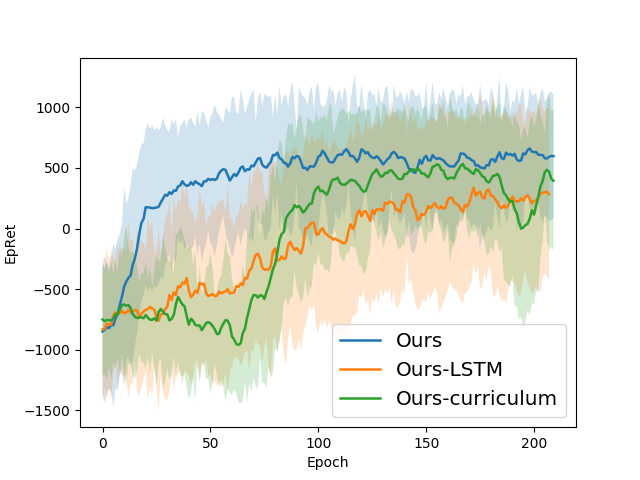}
	}
	\caption{(a) The learning curve of three DRL-based approaches. (b) The learning curve for the ablation study.} 
	\label{fig4:subfig}
\end{figure}

We first compare these approaches in random scenarios. Fig.~\ref{fig4:subfig:a} shows the learning curve of these approaches. It can be seen that the average reward obtained by ours is higher than others and the converging speed of our approach is faster than others, which suggests that DRQN provides a better way to record historical information to handle partial observation as expected.

Table~\ref{table2} shows the performance metrics of these approaches in random scenarios, where metrics are calculated from the averaging results of 200 runs of scenarios. Ours outperforms others in random scenarios. In particular, the success rate (SR) and the expected return (ER) of ours are much better than others. Moreover, generated trajectories of ours are also smoother as it has the lowest average angular velocity change (AAVC). Meanwhile, the mean length of these generated trajectories is similar to others as the reach time (RT) is similar to others.

We also test the performance of 3D obstacle avoidance for all three approaches on other simulation scenarios, i.e., office-like scenarios and coffee-like scenarios. 
Notice that, we only use random scenarios to train the networks in these DRL-based approaches. We want to show that the generated policy from our approach can also perform well in unseen scenarios.
In specific, Table~\ref{table3} shows the success rate (SR) for these approaches in 200 runs of both office-like and coffee-like scenarios.
Ours has the highest SR in all testing scenarios, which 
indicates that our approach can generalize well to new scenarios.

\begin{table}[b] 
	\caption{Success rate (SR) of three approaches in testing scenarios}  
	\begin{center}
		\begin{tabular}{llll}
			\hline\noalign{\smallskip}  
			Method & Random  & Office-like & Coffee-like \\  
			\hline\noalign{\smallskip}  
			Prior work  & 76.5 & 38.5 & 61.5 \\  
			PPO & 57.5 & 56.0 & 60.3 \\  
			Ours & \textbf{92} & \textbf{88.5} & \textbf{81.5} \\
			\hline\noalign{\smallskip} 
		\end{tabular}  
		\label{table3}
	\end{center}
\end{table}

\subsubsection{Ablation Study and Analysis}

We also conduct an ablation study to illustrate the effect of LSTM units and curriculum learning. 
DRQN without LSTM units is equivalent to curricular 1-frame DQN, i.e., the DQN-based approach using curriculum learning and taking one frame of local costmap as input.
Fig.~\ref{fig4:subfig:b} shows the learning curve for the ablation study and Table~\ref{table4} shows corresponding performance metrics in 200 runs of random scenarios.

The experimental results show that both LSTM units and curriculum learning play an important role in the performance of curricular DRQN.
In specific, the training process can be accelerated and stabilized by curriculum learning. 
Notice that, the model needs to learn how to navigate in various environments and it is more challenging for environments with dense obstacles and long-range navigation tasks. 
Clearly, the collision avoidance abilities learned in simple navigation environments can help the model to navigate in more challenging environments.
Then we can use curriculum learning to train the model in multiple navigation environments with different levels of difficulty and transfer the knowledge learned from simple environments to difficult ones.
Fig.~\ref{fig4:subfig:b} and Table~\ref{table4} show that curriculum learning can stabilize and speed up the training.
LSTM units can help the networks to handle historical information for partial observation.
Notice that, curricular 1-frame DQN performs better on AAVC and generates smoother trajectories. Because without the ability to record historical information the approach provides a weaker intention to avoid obstacles.

\begin{table}[b]
	\caption{Metrics for the ablation study} 
	\begin{center}
		\begin{tabular}{lllll}
			\hline\noalign{\smallskip}  
			Method & SR & ER & RT & AAVC \\  
			\hline\noalign{\smallskip}  
			Ours  & \textbf{92} & \textbf{715} & \textbf{6.78} & 0.43 \\  
			-- LSTM & 72.5 & 482 & 7.68 & \textbf{0.38} \\  
			-- Curriculum & 86.5 & 610 & 7.84 & 0.53 \\
			\hline\noalign{\smallskip} 
		\end{tabular}  	 
		\label{table4}  
	\end{center}
\end{table}

\subsection{Experiments in Real World}

\begin{figure}[t]
	\centerline{\includegraphics[width=1.0\linewidth]{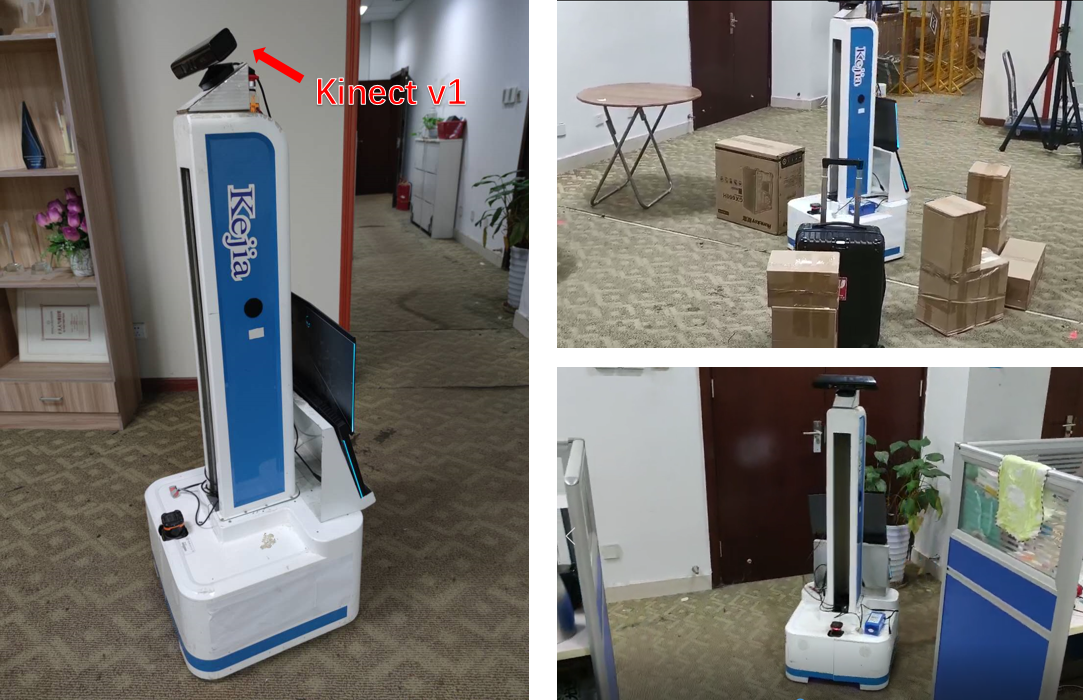}}
	\caption{Our mobile robot platform (left). A testing environment with static obstacles (upper right) and an office scenario (bottom right).} 
	\label{fig5}
\end{figure}

We also deploy the trained model to a real robot to perform 3D obstacle avoidance in the real world. 
As shown in Fig.~\ref{fig5}, the robot is based on a differential platform with a Kinect v1 depth sensor.  
A Hokuyo UTM-30LX scanning laser rangefinder is used to locate the robot in the environment. Notice that, this laser sensor is not necessary, which can be replaced by other equipment for the localization. A laptop with an i7-8750H CPU and an NVIDIA 1060 GPU is used as the computing platform.

In experiments, we use a particle filter-based state estimator to provide the real-time position and velocity of the robot.
From the depth information provided by Kinect, we construct the local costmap,
that has the fixed size $ 6.0 \text{ m } \times 6.0 \text{ m}$ and the resolution $0.1 \text{ m}$.

We use paper boxes as scattered obstacles to create cluttered scenarios for testing.
As shown in Fig.~\ref{fig5}, our robot can adjust its direction and go through obstacles with no collision. We also test the 3D obstacle avoidance ability of our robot in a real office environment.
The demonstration video for both simulation and real-world experiments can be found at
\url{https://youtu.be/_Y9l6hEopyk}.

\section{Conclusions and future work}\label{conclusions}

In this paper, we proposed a map-based end-to-end DRL method for 3D obstacle avoidance with partial observation, which is applied to achieve obstacle avoidance in complex indoor environments only using a depth camera with a narrow field of view. 
The approach is based on double
DRQN and using curriculum learning to accelerate and stabilize the training process. 
Both qualitative and quantitative experiments on simulation environments show that our approach is efficient and outperforms other existing approaches.
We also show that the trained model can be easily deployed to a real robot to perform 3D collision avoidance in its navigation.

In the future, we will try to use more 3D information rather than compressing as a top view to excute obstacle avoidance. Furthermore, we will investigate how to combine multiple sensors to make up for the shortcomings of single sensors. We also want to use our approach on MindSpore~\footnote{https://www.mindspore.cn/}, which is a new deep learning computing framework.

\bibliographystyle{IEEEtran}
\bibliography{IEEEabrv,my}

\end{document}